\documentclass{article}

\usepackage{arxiv}
\usepackage[utf8]{inputenc} 
\usepackage[T1]{fontenc}    
\usepackage{multirow, multicol}
\usepackage{adjustbox}
\usepackage{hyperref}       
\usepackage{url}            
\usepackage{booktabs}       
\usepackage{amsfonts}       
\usepackage{nicefrac}       
\usepackage{microtype}      
\usepackage{lipsum}		
\usepackage{graphicx}
\usepackage{natbib}
\usepackage{doi}
\usepackage{array}
\usepackage{animate}
\usepackage{amsmath,amssymb}
\usepackage{color}

\newcommand{\beginsupplement}{%
        \setcounter{table}{0}
        \renewcommand{\thetable}{S\arabic{table}}%
        \setcounter{figure}{0}
        \renewcommand{\thefigure}{S\arabic{figure}}%
     }

\title{Temporal-Aware Refinement for Video-based Human Pose and Shape Recovery}

\author{ \href{https://orcid.org/0009-0007-3586-6006}{\includegraphics[scale=0.06]{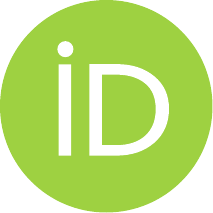}\hspace{1mm}Ming Chen$^*$} \\
	Kuaishou Technology Co., Ltd.\\
	\texttt{chenming09@kuaishou.com} \\
	\And
	\href{https://orcid.org/0009-0002-0564-0915}{\includegraphics[scale=0.06]{orcid.pdf}\hspace{1mm}Yan Zhou$^*$} \\
	Kuaishou Technology Co., Ltd. \\
	\texttt{zhouyan03@kuaishou.com} \\
	\And
         \href{https://orcid.org/0009-0002-0589-5264}{\includegraphics[scale=0.06]{orcid.pdf}\hspace{1mm}Weihua Jian} \\
	Kuaishou Technology Co., Ltd. \\
	\texttt{jianweihua@kuaishou.com} \\
        \And
         \href{https://orcid.org/0000-0001-7225-565X}{\includegraphics[scale=0.06]{orcid.pdf}\hspace{1mm}Pengfei Wan} \\
	Kuaishou Technology Co., Ltd. \\
	\texttt{wanpengfei@kuaishou.com} \\
        \And
        \href{https://orcid.org/0000-0003-2987-9672}{\includegraphics[scale=0.06]{orcid.pdf}\hspace{1mm}Zhongyuan Wang} \\
	Kuaishou Technology Co., Ltd. \\
	\texttt{wangzhongyuan@kuaishou.com} \\
}

\begin{document}
\maketitle
\def\thefootnote{*}\footnotetext{Equal contributions}

\begin{abstract}
Though significant progress in human pose and shape recovery from monocular RGB images has been made in recent years, obtaining 3D human motion with high accuracy and temporal consistency from videos remains challenging. Existing video-based methods tend to reconstruct human motion from global image features, which lack detailed representation capability and limit the reconstruction accuracy. In this paper, we propose a Temporal-Aware Refining Network (TAR), to synchronously explore temporal-aware global and local image features for accurate pose and shape recovery.  First, a global transformer encoder is introduced to obtain temporal global features from static feature sequences. Second, a bidirectional ConvGRU network takes the sequence of high-resolution feature maps as input, and outputs temporal local feature maps that maintain high resolution and capture the local motion of the human body. Finally, a recurrent refinement module iteratively updates estimated SMPL parameters by leveraging both global and local temporal information to achieve accurate and smooth results. Extensive experiments demonstrate that our TAR obtains more accurate results than previous state-of-the-art methods on popular benchmarks, i.e., 3DPW, MPI-INF-3DHP, and Human3.6M.
\end{abstract}


\begin{figure}[htb]
    \centering
    \animategraphics[autoplay,loop, width=0.8\textwidth]{15}{./figures/compare_sample/compare_sample-}{100}{144}
    \caption{Previous methods TCMR\cite{choi2021beyond}(white) and GLoT\cite{shen2023global}(purple) solely focus on leveraging global image features, which leads to inaccurate and misaligned meshes. In contrast, TAR(brown) modeling both global and local temporal information to capture more accurate human motion. \textit{This is a video figure that is best viewed by Adobe Reader.}}
    \label{fig:compare-gif}
\end{figure}

\section{Introduction}
\label{sec:intro}

Monocular RGB image-based human pose and shape recovery is crucial in AR/VR, motion analysis, computer games, and human-computer interaction. In recent years, various image-based methods\cite{kanazawa2018end}\cite{kolotouros2019learning}\cite{kocabas2021pare}\cite{wang2023refit} are proposed to tackle recovering human pose and shape from RGB images. Some methods directly estimate the parameters of a parametric model (e.g., SMPL\cite{SMPL:2015}), while others reconstruct the vertex coordinates of the human body surface. 

Despite impressive improvements in image-based methods, capturing temporal-consistent human motion remains challenging. Video-based methods have been proposed to address this issue. Previous video-based methods typically design a network to model the long-term or local-temporal relations among global feature sequences of low-resolution extracted by a pretrained backbone\cite{kolotouros2019learning}. For instance, TCMR\cite{choi2021beyond} and GLoT\cite{shen2023global} consist of a GRU-based and Transformer-based temporal encoder, respectively. Although video-based methods have made progress improving intra-frame accuracy and inter-frame consistency, they still struggle to achieve the high accuracy of image-based counterparts. Most previous video-based methods generally focus on modeling global features processed by the pooling operation of CNN backbones, neglecting critical local image features that contain detailed information on human pose and shape. Besides, these methods apply weakly-perspective assumption used in \cite{kanazawa2018end}, which is not suitable in many real scenarios and leads to undesirable results. 

To address the issues mentioned above, we propose the Temporal-Aware Refining Network (TAR) that utilizes temporal-aware global and local image features with a iterative regression module. Our method consists of a Global Temporal Encoder, a Local Temporal Encoder, and a Recurrent Refinement Module. The Global Temporal Encoder responses for temporal global information encoding with a 4-layer transformer encoder. The Local Temporal Encoder comprises a bidirectional ConvGRUs and focuses on modeling the dynamics of local image features. To fully utilize temporal features of different semantic levels, we propose a Recurrent Refinement Module that iteratively integrates local and global features to update predictions. Besides, we leverage full-image perspective camera model in the training phase to overcome the shortage of weakly perspective assumption\cite{li2022cliff}. With these designs, TAR outperforms previous state-of-the-art methods on public 3DPW, MPI-INF-3DHP, and Human3.6M datasets. In summary, our contributions are as followed:
\begin{itemize}
    \item We propose a Tempora-Aware Refining Network (TAR) for recovering 3D human mesh from video. TAR collaboratively exploits temporal global and local image features with Global and Local Temporal Encoders, achieving accurate and temporally consistent results.
    \item We propose a Recurrent Refinement Module to iteratively leverage temporal-aware global and local image features to refine predicted SMPL parameters.
    \item We conduct extensive experiments on three public datasets\cite{von2018recovering}\cite{ionescu2013human3}\cite{mehta2017monocular}. The results show that TAR surpasses the performance of previous methods. In addition, we also verify the generalization ability of TAR on the recent challenging dataset EMDB\cite{kaufmann2023emdb}.
    
\end{itemize}

\section{Related Work}
\label{sec:related}

\subsection{Image-based human pose and shape recovery.}
3D human mesh recovery aims to reconstruct 3D human pose and shape from monocular images. Previous image-based methods can be divided into parametric methods and non-parametric methods. The first type estimates parameters of parametric human model (e.g., SMPL\cite{SMPL:2015}), while the second type reconstructs mesh vertices of human body. Early works\cite{kanazawa2018end}\cite{kolotouros2019learning}\cite{kocabas2021spec} directly predict SMPL pose and shape parameters from global image features extracted by pretrained CNNs\cite{he2016deep}. However, they struggle to generate accurate results due to the limited representation capability of global features. Recent works\cite{kocabas2021pare}\cite{zhang2021pymaf}\cite{wang2023refit} leverage local image features to complement the representation ability of global features and achieve better accuracy. Several works\cite{li2021hybrik}\cite{li2023niki}\cite{song2020human}\cite{georgakis2020hierarchical}\cite{ma20233d} employ intermediate representations and multiple stages to simplify parameters estimation. For example, HybrIK\cite{li2021hybrik} conducts swing-twist decomposition to joint rotations, breaking the task into two phases: detecting 3D keypoints and joint twists from images and computing poses via inverse kinematics. Other works\cite{choi2020pose2mesh}\cite{moon2020pose2pose}\cite{kolotouros2019convolutional}\cite{lin2021mesh}\cite{lin2021end}\cite{cho2022cross} directly predict 3D coordinates of mesh vertices from images\cite{kolotouros2019convolutional}\cite{lin2021mesh}\cite{lin2021end}\cite{cho2022cross} or 2D keypoints\cite{choi2020pose2mesh}\cite{moon2020pose2pose} to overcome the limited pose and shape representation space of SMPL model. Although image-based methods have made remarkable progress in accuracy, generating temporally consistent human motion from videos remains an open problem.

\subsection{Video-based human pose and shape recovery.}
Video-based methods aim to produce accurate and temporal-consistent human mesh from sequential video frames. Most methods\cite{kocabas2020vibe}\cite{choi2021beyond}\cite{wei2022capturing}\cite{shen2023global}\cite{luo20203d}
use a pretrained CNN\cite{kolotouros2019learning} to first extract static global features from image sequences and build temporal encoder such as attention\cite{wei2022capturing}, GRUs\cite{kocabas2020vibe}\cite{choi2021beyond} and Transformers\cite{shen2023global}, to predict SMPL parameters. Other methods\cite{wan2021encoder}\cite{yang2023capturing} train a Vision Transformer\cite{dosovitskiy2020image} to extract fine-grained image features and exploit spatial-temporal relations from input sequences. MAED\cite{wan2021encoder} takes the image patch sequence as input and uses a spatial-temporal Transformer framework to predict SMPL parameters. INT\cite{yang2023capturing} first trains a ViT\cite{dosovitskiy2020image} to learn disentangled pose tokens from images and uses multiple temporal Transformers to capture the motion of each token. Despite the promising results of video-based methods, these mthods only consider the global image features while discarding detailed spatial information, leading to trade-off between accuracy and smoothness. In contrast, our method simultaneously utilize temporal-aware local and global image features, achieving better per-frame accuracy and temporal consistency.

\begin{figure}[htb]
\centering
\includegraphics[height=5.0cm]{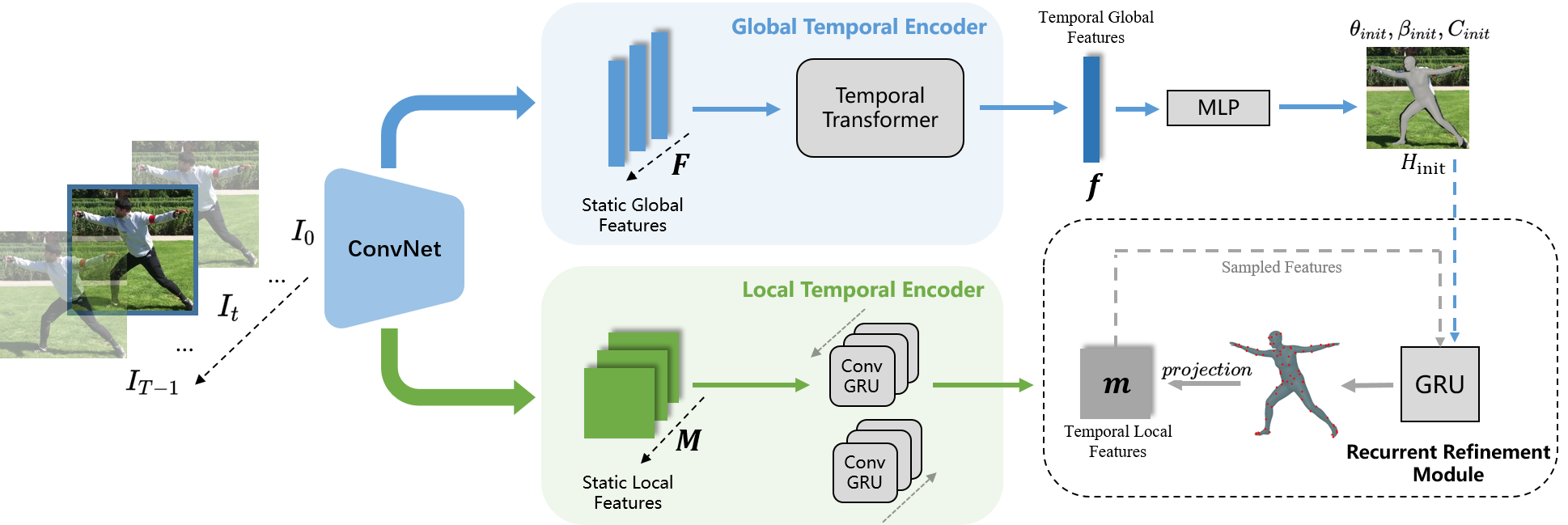}
\caption{An overview of the proposed TAR. Given a sequence of images, static global features and local features are first extracted by a pretrained CNN\cite{he2016deep}\cite{wang2020deep}. Then a transformer-based\cite{vaswani2017attention} Global Temporal Encoder takes the static global features $\mathbf{F}$ as input and obtains temporal global features $\mathbf{f}$ of the mid-frame, which is further converted to initial SMPL parameters and hidden states $\mathbf{H}_{init}$ by a MLP network. Parallelly, a Local Temporal Encoder composed of a bidirectional ConvGRU\cite{siam2017convolutional} network extracts temporal local features $\mathbf{m}$ from the static local features $\mathbf{M}$. Last, Recurrent Refinement Module initialized by $\mathbf{H}_{init}$ utilizes $\textbf{f}$ and $\textbf{m}$ to iteratively update the estimated SMPL parameters.}
\label{fig:framework}
\end{figure}

\section{Method}
\label{sec:method}

Figure ~\ref{fig:framework} shows an overview of our Temporal-Aware Refining Network (TAR). TAR comprises three components: (1) A Global Temporal Encoder (GTE), (2) a Local Temporal Encoder (LTE) and (3) a Recurrent Refinement Module (RRM). Given an image sequence $\mathbf{I}$ of length $T$, $\mathbf{I}=\left \{ I_t \right \}^{T-1}_{0}$, we first use a pretrained CNN\cite{he2016deep}\cite{siam2017convolutional} to extract the image features. Differing from previous methods, in addition to the global features $\mathbf{F}=\left \{ f_t \right \}^{T-1}_{0}$, we extract high-resolution (e,g., $64\times64$) local feature maps $\mathbf{M}=\left \{ m_t \right \}^{T-1}_{0}$ to capture the detailed spatial information. Static global and local features are used as input to produce temporal-aware global and local representations, denoted by $\mathbf{f}$ and $\mathbf{m}$, respectively, by GTE and LTE. Finally, both types of features are fed to the RRM, which iteratively updates the estimated SMPL parameters of the mid-frame in the input sequence. We elaborate on each component of TAR as follows.

\begin{figure}[htb]
\centering
\includegraphics[height=6.0cm]{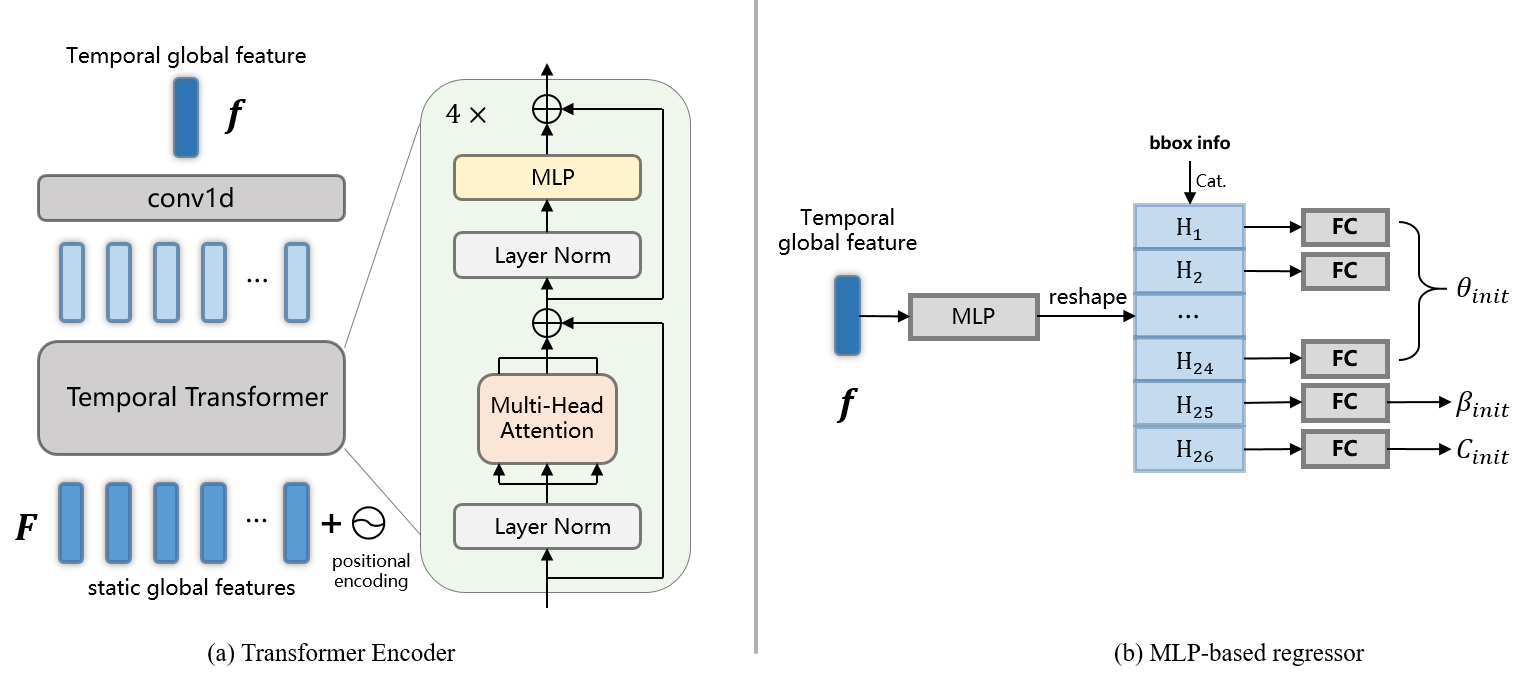}
\caption{(a) Transformer encoder structure of Global Temporal Encoder and weighted average layer implemented with 1D convolution. (b) MLP-based regressor with 26 hidden states and bounding box information vector.}
\label{fig:gte}
\end{figure}

\subsection{Global Temporal Encoder (GTE)}
GTE involves two components, i.e. a temporal transformer encoder and an MLP-based regressor proposed by \cite{wang2023refit}. The transformer encoder processes the static global features to obtain temporal-aware global features of the mid-frame. The regressor first converts the features into disentangled hidden states, which are latter used to regress the initial SMPL parameters.
\subsubsection{Transformer Encoder.} Transformers\cite{vaswani2017attention} have shown a powerful capability to capture long-range global dependency of sequential data thanks to the self-attention mechanism. As shown in Figure \ref{fig:gte}(a), our GTE applies a 4-layer transformer encoder to learn the temporal consistency in human motion. Then a weighted average layer shrinks the entire sequence into a temporal global feature vector, which integrates historical and future information within the sequence.
\subsubsection{Initial Regressor.} After performing global temporal encodeing, we estimate the SMPL\cite{SMPL:2015} parameters  $\Phi_{init} = \left \{ \theta_{init}, \beta_{init}, C_{init} \right \}, \theta_{init} \in \mathbb{R}^{24\times3},\beta_{init}\in \mathbb{R}^{10}, C_{init}\in \mathbb{R}^{3}$ from the integrated temporal global features. $\theta$ and $\beta$ are pose and shape parameters that control the joint rotations and mesh shapes in SMPL, and $C$ represents camera parameters\cite{li2022cliff} which project the 3D coordinates onto the full-image space. Instead of using the common iterative regressor proposed by HMR\cite{kanazawa2018end}, we first convert the global features into 26 hidden states corresponding to 23 SMPL joint rotations, global orientation, shape coefficient and camera parameters. Together with bounding box information vector\cite{li2022cliff}, all hidden states are sent to an MLP-based regressor to estimate SMPL parameters as shown in Figure \ref{fig:gte}(b). The hidden states and estimated parameters are subsequently used to initialize the Recurrent Refinement Module. 

\begin{figure}[htb]
\centering
\includegraphics[height=5.0cm]{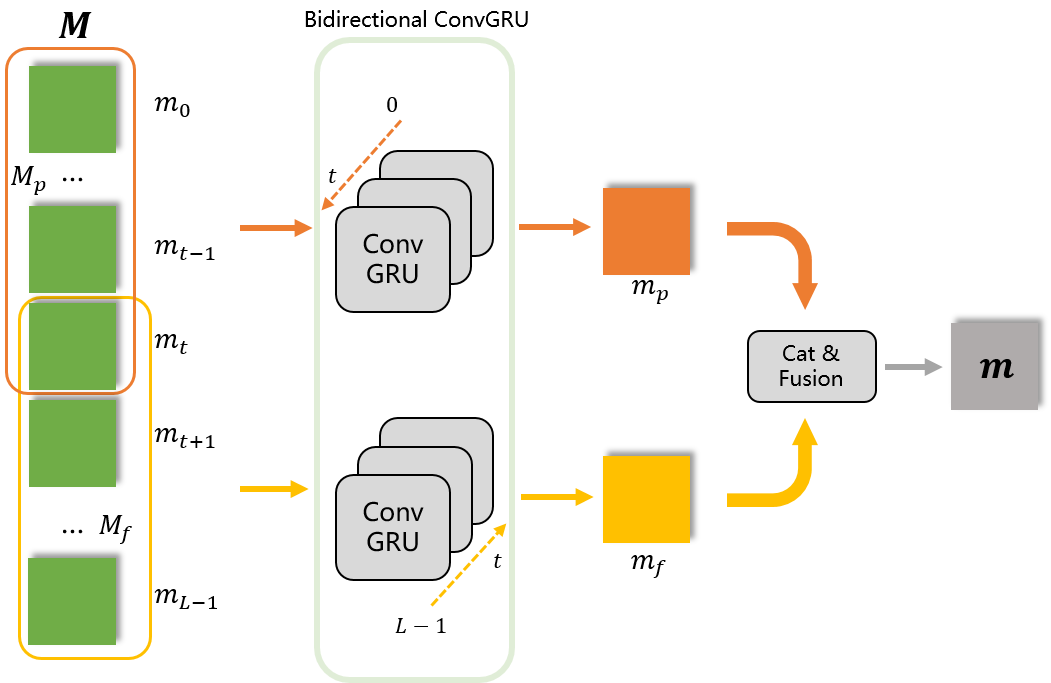}
\caption{Local Temporal Encoder built with bidirectional ConvGRU. $\mathbf{M}$ is a sequence local features with the shape of $L\times D\times h \times w$, where $L$ is the size of local temporal window, $D$ and $(h, w)$ is the channel dimension and spatial resolution of local feature maps. $\mathbf{m}_f$, $\mathbf{m}_p$ are the outputs of history and future branch, respectively. $\mathbf{m}$ is the final temporal-aware local features extracted by LTE.}
\label{fig:lte}
\end{figure}

\subsection{Local Temporal Encoder}
The GTE integrates global features while disregarding essential local information in images, which has been proved beneficial for 3D accuracy and 2D alignment. As compensation, LTE exploits high-resolution representations with temporal awareness from static local feature maps to capture the local motion of human body.

\subsubsection{Convolutional Gated Recurrent Units.} ConvGRUs are wildly used in video segmentation tasks\cite{siam2017convolutional}\cite{lin2022robust} to exploit temporal relations in high-resolution images. Inspired by this, we build the Local Temporal Encoder with a bidirectional ConvGRU network to leverage both future and historical temporal information from local features. As shown in Figure \ref{fig:lte}, the sequence of local features $\mathbf{M}=\left \{ m_t \right \}^{L-1}_{0}, m\in \mathbb{R}^{D\times h\times w}$ is first split into two parts $\mathbf{M}_p = \left \{ m_t \right \}^{\tau}_{0}$ and $\mathbf{M}_f = \left \{ m_t \right \}^{\tau}_{L-1}$, where $\tau=\left \lfloor (L-1)/2 \right \rfloor $ is the index of the mid-frame. Then, the bidirectional ConvGRU computes the hidden maps $\mathbf{m}_p$, $\mathbf{m}_f$ from both sub-sequences. In each ConvGRU block, the computation process can be writen as 
\begin{align}
    & Z_t = \sigma(W_z * m_t + U_z * h_{t-1}) \\
    & R_t = \sigma(W_r * m_t + U_z * h_{t-1}) \\
    & \tilde{h}_t = tanh(W * m_t + U * (R_t \circ h_{t-1})) \\
    & h_t = (1 - Z_t) \circ h_{t-1} + Z_t \circ \tilde{h}_t,
\end{align}
where $h_t, h_{t-1}$ represent hidden maps of ConvGRU at different time steps, and $W$, $U$ are learnable weights. The hidden maps $h$ have the same shape with input local features $m_t$ and are initialized with zeros in both branches. Last, a convolutional layer concatenates and projects the hidden maps $\mathbf{m}_f$, $\mathbf{m}_p$ into the temporal-aware local features $\mathbf{m}$. 

It is worth noting that, the size of window $L$ of LTE can be different from length $T$ of input sequence, and we empirically find that LTE benefits from smaller $L$ than $T$. We assume the reason is that ConvGRU is more suitable for capturing local motion information in a short period instead of long-range modeling. The experiments in Session \ref{sec:exp} verify the impact of $L$.

\begin{figure}[htb]
\centering
\includegraphics[height=5.0cm]{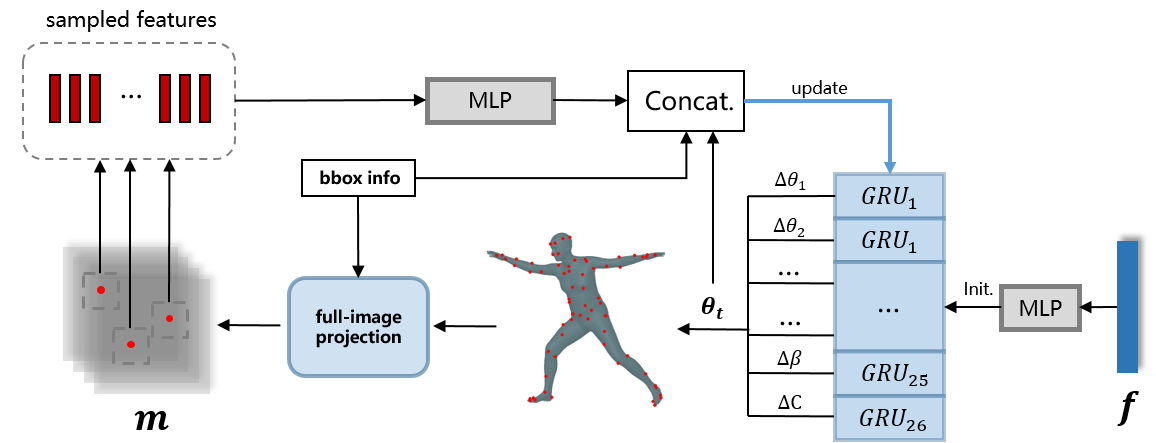}
\caption{The proposed Recurrent Refinement Module. It contains 26 disentangled GRUs to update SMPL parameters with local features and bounding box information.}
\label{fig:rrm}
\end{figure}

\subsection{Recurrent Refinement Module}
Previous video-based methods\cite{kocabas2020vibe}\cite{choi2021beyond}\cite{wei2022capturing}\cite{shen2023global} generally apply a regressor to estimate SMPL parameters from temporal global features without utilizing local image details. Inspired by recent image-based methods\cite{zhang2021pymaf}\cite{wang2023refit}, we propose Recurrent Refinement Module (RRM) to update estimated parameters with collaborative usage of temporal local and global features. Specifically, we build 26 disentangled GRUs to update SMPL poses, shape and camera parameters. In each iteration $l$, the GRUs take feedback signal $\mathbf{s}_{l}$ as input, and predicts an update step $\Delta \Phi_{l}$, which is added to produce the next estimation: $\Phi_{l+1} = \Phi_{l} + \Delta\Phi_{l}$. 
\subsubsection{Feedback Signal.} Given the initial estimate $\Phi_{init}$ from GTE, we first obtain the SMPL mesh via forward kinematics and linear blend shapes\cite{SMPL:2015}. Following \cite{wang2023refit}, $K$ markers $\mathbf{J}=\left \{ J_k=(x_k, y_k, z_k) \right \} ^{k=26}_{k=1}$ are sampled from the SMPL mesh and projected and transformed to bounding box space to get $K$ keypoints $\mathbf{j}=\left \{ j_k=(u_k, v_k) \right \}^{k=K}_{k=1}$. As proposed in CLIFF\cite{li2022cliff}, we conduct the more reasonable full-image projection instead of the widely used weakly-perspective projection:
\begin{align}
    & j^{full}_k = \Pi(J_k + t^{full}) \\
    & j_k = (j^{full}_k - c^{bbox}) / s^{bbox}
\end{align}
where $t^{full}$ is the translation with respect to the optical center of the full image, which is estimated from camera parameters $C$\cite{li2022cliff}. $c^{bbox}$ and $s^{bbox}$ are the center and size of the bounding box from which the crop is obtained.
For each $j_k$, we take the feature values inside a window centered at $j_k$ as
\begin{equation}
    s_k = \left \{g(j) \in \mathbf{m} | \left \| j - j_k  \right \| \leq r  \right \}
\end{equation}
where we set $r=3$ pixels as the radius of the window. We then concatenate sampled feature values of all the markers, along with current estimate $\Phi_{l}$ and bounding box information vector $v^{cliff}$ proposed in \cite{li2022cliff} to form the feedback signal:
\begin{equation}
    \mathbf{s}_l=\left [ s_1, \dots, s_K, \Phi_l, v^{cliff} \right ] 
\end{equation}

\subsubsection{Update Step.} Once the feedback signal $\mathbf{s}_l$ is obtained, the hidden states of all GRUs are updated parallelly:
\begin{equation}
    h^{l+1}_{n} \leftarrow (h^l_n, \mathbf{s}_l), n=1,\dots,26
\end{equation}
and 26 individual regressors map the hidden states to update step $\Delta\Phi_l$. There are two major differences between RRM and the feedback module in \cite{wang2023refit}. First, the hidden states of GRUs are initialized with global temporal features, which is zero-initialized in \cite{wang2023refit} instead. Second, our RRM sampled from temporal-aware local features, capturing more temporal dynamics of human body compared to the static features used in\cite{wang2023refit}. These two differences promote our method to capture smoother and more accurate human motion.

\subsection{Loss Functions}

Following previous methods\cite{choi2021beyond}\cite{wei2022capturing}, we impose $L_2$ loss between the estimated and ground truth SMPL parameters and 2D/3D joint coordinates to supervise the initial results of all update stages in GTE and RRM:
\begin{align}
    & L_l = \lambda_{2D}L_{2D}^l + \lambda_{3D}L_{3D}^l + \lambda_{SMPL}L_{SMPL}^l \\
    & L_{2D}^l = \left \| \hat{\mathbf{J}}_{2D} - \mathbf{J}_{2D}^l \right \| ^2_F \\
    & L_{3D}^l = \left \| \hat{\mathbf{J}}_{3D} - \mathbf{J}_{3D}^l \right \| ^2_F \\
    & L_{SMPL}^l = \left \| \hat{\mathbf{\theta}} - \mathbf{\theta}^l \right \| ^2_F + \left \| \hat{\mathbf{\beta}} - \mathbf{\beta}^l \right \| ^2_F
\end{align}
the hat operator denotes the ground truth of that variable.
The final loss is a weighted sum of the loss at each iterative update and initial state
\begin{equation}
    L = \sum^{L}_{l=0}\gamma^{L-l}L_l
\end{equation}
where we set $\gamma$=0.85 and $L$=5 for all experiments following \cite{wang2023refit}. To be noted, our model only estimate the parameters at the mid-frame of the sequence.

\section{Experiments}
\label{sec:exp}
We first introduce the evaluation metrics and datasets. Next, we report the evaluation results of our method and comparison with previous methods. Finally, we provide ablation studies to prove the effectiveness of the proposed method.

\subsubsection{Evaluation Metrics.} Following previous methods\cite{choi2021beyond}\cite{wei2022capturing}\cite{shen2023global}, we report results in the metrics of intra-frame accuracy temporal consistency. For intra-frame metrics, we adopt Mean Per Joint Position Error (MPJPE), Procrustes-aligned MPJPE (PA-MPJPE), and Mean Per Vertex Position Error (MPVPE). For temporal consistency, we report acceleration error (ACCEL) to verify inter-frame smooth.

\subsubsection{Datasets.} We adopt mixed 2D and 3D datasets for training. For 3D datasets, we use 3DPW\cite{von2018recovering}, Human3.6M\cite{ionescu2013human3} and MPI-INF-3DHP\cite{mehta2017monocular}, which contain annotations of SMPL parameters or 3D joints. We do not use Mosh\cite{loper2014mosh} data of Human3.6M, instead we use the pseudo ground truth from NeuralAnnot\cite{moon2022neuralannot}. For 2D datasets, we use COCO\cite{lin2014microsoft} and MPII\cite{andriluka20142d}, which contain 2D joint annotations and pseudo SMPL parameters fitted by EFT\cite{joo2021exemplar}. Since COCO and MPII are both image-based datasets, we repeat each image $T$ times to form a sequence for training. For evaluation, we use the test set of 3DPW, Human3.6M and MPI-INF-3DHP. Additionally, we evaluate our method on a new challenging 3D dataset EMBD\cite{kaufmann2023emdb} to demonstrate the generalization performance.

\subsection{Comparison with state-of-the-art methods}

\begin{table}[htb]\small
\centering
\caption{Evaluation of state-of-the-art methods on 3DPW, MPI-INF-3DHP, and Human3.6M datasets. All Methods use 3DPW training set in training phase. $\dagger$ and $\ast $ represent using HRNet\cite{wang2020deep} and ResNet50\cite{he2016deep} backbone, respectively. \textbf{Bold}: best; \underline{Underline}: second best.}
\label{tab:main}
\resizebox{\columnwidth}{!}{%
\renewcommand\arraystretch{1.6}
\begin{tabular}{cccccccccccc}
\toprule[1.5pt]
\multicolumn{1}{c|}{\multirow{2}{*}{Method}} & \multicolumn{4}{c|}{3DPW} & \multicolumn{3}{c|}{MPI-INF-3DHP} & \multicolumn{3}{c|}{Human3.6M} & \multirow{2}{*}{\begin{tabular}[c]{@{}c@{}}Number of \\ input frames\end{tabular}} \\
\multicolumn{1}{c|}{} & MPJPE$\downarrow$ & PA-MPJPE$\downarrow$ & PVE$\downarrow$ & \multicolumn{1}{c|}{ACCEL$\downarrow$} & MPJPE$\downarrow$ & PA-MPJPE$\downarrow$ & \multicolumn{1}{c|}{ACCEL$\downarrow$} & MPJPE$\downarrow$ & PA-MPJPE$\downarrow$ & \multicolumn{1}{c|}{ACCEL$\downarrow$} &  \\ \midrule
\multicolumn{1}{c|}{VIBE\cite{kocabas2020vibe}} & 91.9 & 57.6 & 99.1 & \multicolumn{1}{c|}{25.4} & 103.9 & 68.9 & \multicolumn{1}{c|}{27.3} & 65.9 & 41.5 & \multicolumn{1}{c|}{18.3} & 16 \\
\multicolumn{1}{c|}{MEVA\cite{luo20203d}} & 86.9 & 54.7 & - & \multicolumn{1}{c|}{11.6} & 96.4 & 65.4 & \multicolumn{1}{c|}{11.1} & 76.0 & 53.2 & \multicolumn{1}{c|}{15.3} & 90 \\
\multicolumn{1}{c|}{TCMR\cite{choi2021beyond}} & 86.5 & 52.7 & 102.9 & \multicolumn{1}{c|}{\underline{7.1}} & 97.6 & 63.5 & \multicolumn{1}{c|}{\underline{8.5}} & 62.3 & 41.1 & \multicolumn{1}{c|}{\underline{5.3}} & 16 \\
\multicolumn{1}{c|}{MAED\cite{wan2021encoder}} & 79.1 & 45.7 & 92.6 & \multicolumn{1}{c|}{17.6} & \textbf{83.6} & \textbf{56.2} & \multicolumn{1}{c|}{-} & 56.4 & 38.7 & \multicolumn{1}{c|}{-} & 64 \\
\multicolumn{1}{c|}{MPS-Net\cite{wei2022capturing}} & 84.3 & 52.1 & 99.7 & \multicolumn{1}{c|}{7.4} & 96.7 & 62.8 & \multicolumn{1}{c|}{9.6} & 69.4 & 47.4 & \multicolumn{1}{c|}{ \textbf{3.6}} & 16 \\
\multicolumn{1}{c|}{INT\cite{yang2023capturing}} & 75.6 & \underline{42.0} & 87.9 & \multicolumn{1}{c|}{16.5} & - & - & \multicolumn{1}{c|}{-} & 54.9 & 38.4 & \multicolumn{1}{c|}{-} & 64 \\
\multicolumn{1}{c|}{GLoT\cite{shen2023global}} & 80.7 & 50.6 & 96.3 & \multicolumn{1}{c|}{\textbf{6.6}} & 93.9 & 61.5 & \multicolumn{1}{c|}{\textbf{7.9}} & 67.0 & 46.3 & \multicolumn{1}{c|}{\textbf{3.6}} & 16 \\ \midrule
\multicolumn{1}{c|}{\textbf{TAR(Ours)$\ast$}} & \underline{68.4} & 45.3 & 83.9 & \multicolumn{1}{c|}{7.9} & 87.5 & 62.7 & \multicolumn{1}{c|}{9.6} & \underline{53.1} & \underline{37.8} & \multicolumn{1}{c|}{5.6} & \textbf{9} \\
\multicolumn{1}{c|}{\textbf{TAR(Ours)$\dagger$}} & \textbf{62.7} & \textbf{40.6} & \textbf{74.4} & \multicolumn{1}{c|}{7.7} & \underline{85.9} & \underline{60.5} & \multicolumn{1}{c|}{9.2} & \textbf{45.6} & \textbf{33.3} & \multicolumn{1}{c|}{5.6} & \textbf{9} \\ \bottomrule[1.5pt]
 &  &  &  &  &  &  &  &  &  &  & 
\end{tabular}%
}
\end{table}

\subsubsection{Video-based methods.} Table \ref{tab:main} compares our method with the state-of-the-art video-based methods on 3DPW, MPI-INF-3DHP and Human3.6M datasets. Our TAR outperforms existing methods by a large margin on Human3.6M and challenging 3DPW in 3D accuracy metrics, and achieve comparable results on MPI-INF-3DHP dataset. Specifically, TAR brings 15.4mm and 9.3mm decrease of MPJPE on 3DPW and Human3.6M. Although previous methods, e.g., MPS-Net\cite{wei2022capturing} and GLoT\cite{shen2023global} achieve slendid inter-frame smoothness by temporal attention over static global features extracted by SPIN\cite{kolotouros2019learning}, they ignore the detailed local features and use the inappropriate weakly-perspective assumption. Due to these drawbacks, these methods fail to perceive subtle motion of human body and produce large error in real scenario. MAED\cite{wan2021encoder} and INT\cite{yang2023capturing} train ViT\cite{dosovitskiy2020image} as the backbone and take use of fine-grained feature maps to achieve more accurate results. Nevertheless, these methods are time-costing and memory-intensive since they require longer input sequence, e.g., $T$=64. Moreover, these they fail to achieve temporal consistent results compared to other methods. In contrast, our method can estimate accurate and smooth human motion from video with a small input sequence $T$=9. The results demonstrate that exploiting temporal local features and the recurrent refinement mechanism can effectively achieve high intra-frame accuracy and inter-frame consistency.

\begin{table}[htb]
\centering
\caption{Evaluation of state-of-the-art methods on 3DPW. All methods do not use 3DPW on training.}
\label{tab:main-pw3d}
\begin{tabular}{lc|cccc}
\toprule[1.5pt]
\multicolumn{2}{c|}{\multirow{2}{*}{Method}} & \multicolumn{4}{c}{3DPW}                 \\
\multicolumn{2}{c|}{}                        & MPJPE$\downarrow$     & PA-MPJPE$\downarrow$ & MPVPE$\downarrow$ & ACCEL$\downarrow$ \\ \midrule
\multirow{7}{*}{image-based}   & HMR\cite{kanazawa2018end}         & 130.0         & 76.7     & -     & 37.4  \\
                               & Pose2Pose\cite{moon2020pose2pose}   & 86.6          & 54.4     & -     & -     \\
                               & Pose2Mesh\cite{choi2020pose2mesh}   & 88.9          & 58.3     & -     & -     \\
                               & SPIN\cite{kolotouros2019learning}        & 96.9          & 59.2     & 116.4 & 29.8  \\
                               & PyMAF\cite{zhang2021pymaf}       & 78.0          & 47.1     & 91.3  & -     \\
                               & PARE\cite{kocabas2021pare}        & 82.0          & 50.9     & 97.9  & -     \\
                               & HybrIK\cite{li2021hybrik}      & 80.0          & 48.8     & 94.5  & -     \\ \midrule
\multirow{7}{*}{video-based}   & HMMR\cite{kanazawa2019learning}        & 116.5         & 72.6     & 139.3 & 15.2  \\
                               & VIBE\cite{kocabas2020vibe}        & 93.5          & 56.5     & 113.4 & 27.1  \\
                               & TCMR\cite{choi2021beyond}        & 95.0          & 56.5     & 111.5 & 7.0   \\
                               & MPS-Net\cite{wei2022capturing}     & 91.6          & 54.0     & 109.6 & 7.5   \\
                               & INT\cite{yang2023capturing}         & 90.0          & 49.7     & 105.1 & 23.5  \\
                               & GLoT\cite{shen2023global}        & 89.9          & 53.5     & 107.8 & \textbf{6.7 }  \\
                               & \textbf{TAR(Ours)}   & \textbf{71.0} & \textbf{46.3}     & \textbf{84.7}  & 7.4   \\ \bottomrule[1.5pt]
\end{tabular}
\end{table}

\subsubsection{Image-based and video-based methods}
We further compare our TAR with the previous image-based and video-based methods on the challenging in-the-wild 3DPW dataset. Notice that all methods did not use the 3DPW training set in the training phases. The evaluation results are demonstrated in Table \ref{tab:main-pw3d}. Our TAR outperforms existing methods on all intra-frame metrics, and also achieve comparable smoothness with state-of-the-art video-based methods. For example, TAR significantly surpasses previous single image-based and video-based methods by 9.0mm and 18.9mm on MPJPE metric. The results confirm that our proposed TAR is capable for capturing accurate and smooth motion from videos.

\subsection{Ablation Study}

 \subsubsection{Recurrent Refinement Module (RRM)}
 TAR collaboratively use temporal global and local features in a iterative manner. Without RRM, only temporal global features are used to estimate final results, i.e., the initial estimate $\Phi_{init}$ in GTE. As shown in Table \ref{tab:abl_rrm}, the intra-frame accuracy dramatically degrades without RRM. The evaluation results verify that utilizing local features is crucial to improving 3D accuracy.

\begin{table}[htb]
\centering
\caption{Ablation study for Recurrent Refinement Module on 3DPW dataset.}
\label{tab:abl_rrm}
\renewcommand\arraystretch{1.3}
\begin{tabular}{c|cccc}
\toprule[1.5pt]
Method  & MPJPE$\downarrow$         & PA-MPJPE$\downarrow$      & PVE$\downarrow$           & ACCEL$\downarrow$        \\ \midrule
w/o RRM & 73.7          & 45.8          & 86.5          & \textbf{6.8} \\
Ours    & \textbf{62.7} & \textbf{40.6} & \textbf{74.4} & 7.7          \\ \bottomrule[1.5pt]
\end{tabular}
\end{table}

\subsubsection{Global Temporal Encoder (GTE) and Local Temporal Encoder (LTE).}
 There are two temporal encoders in our TAR, i.e., Global Temporal Encoder and Local Temporal Encoder, to respectively capture temporal-aware global and local features from given input sequences. The evaluation results of temporal encoders are in Table \ref{tab:abl_encoder}. By removing LTE and GTE, the performance decreases in both intra-frame and inter-frame metrics, especially ACCEL. The GTE captures the motion by integrating global features, which is essential to the temporal perception ability. However, isolately using GTE is not sufficient for producing accurate and smooth results. The reason is that, in Recurrent Refining Module, local features are iteratively used to update the hidden states initialized by temporal global features. Without LTE, the local features are temporal-agnostic and will degrade the temporal awareness of global features from GTE. The results prove that both GTE and LTE are essential to our TAR.

\begin{table}[htb]
\centering
\caption{Ablation study for Global and Local Temporal Encoders on 3DPW dataset.}
\label{tab:abl_encoder}
\renewcommand\arraystretch{1.3}
\begin{tabular}{c|cccc}
\toprule[1.5pt]
Method     & MPJPE$\downarrow$         & PA-MPJPE$\downarrow$      & PVE$\downarrow$           & ACCEL$\downarrow$        \\ \midrule
only GTE   & 66.7          & 41.8          & 78.5          & 16.0         \\
only LTE   & 65.4          & 41.1          & 76.4          & 19.7         \\
Both(Ours) & \textbf{62.7} & \textbf{40.6} & \textbf{74.4} & \textbf{7.7} \\ \bottomrule[1.5pt]
\end{tabular}
\end{table}

\subsubsection{Lengths of nearby frames of LTE.} As illustrated in Section \ref{sec:method}, the size of window in LTE can be different from the input sequence. We evaluate the impact of different lengths of nearby frames of LTE. As shown in Table \ref{tab:abl_length}, we discover that the model achieves the best results when $L$=5. Although the estimation results gain better temporal consistency when the sequence gets longer, the 3D accuracy also degrades. We assume one probable reason is that ConvGRUs can effectively capture short-term local temporal features, while lacks sufficient modeling ability. Besides, the evaluation results reveal that even small sequences of local features lead to conspicuous temporal-consistency improvement, which proves the significance of temporal local features in TAR.

\begin{table}[htb]
\centering
\caption{Ablation study of different lengths of nearby frames of LTE. The length of 5 means 2 frames before and after the current frame.}
\label{tab:abl_length}
\renewcommand\arraystretch{1.3}
\begin{tabular}{c|cccc}
\toprule[1.5pt]
Method       & MPJPE$\downarrow$         & PA-MPJPE$\downarrow$      & PVE$\downarrow$           & ACCEL$\downarrow$        \\ \midrule
L=3          & 64.9          & 42.1          & 77.4          & 9.0          \\
\textbf{L=5} & \textbf{62.7} & \textbf{40.6} & \textbf{74.4} & 7.7          \\
L=7          & 64.7          & 42.1          & 77.3          & \textbf{7.2} \\
L=9          & 66.4          & 42.5          & 79.1          & 7.3          \\ \bottomrule[1.5pt]
\end{tabular}
\end{table}

\subsection{Additional Experiment}

\subsubsection{Evaluation on EMDB dataset}
EMDB\cite{kaufmann2023emdb} is a recently proposed electromagnetic dataset of 3D human pose and shape in the wild. It contains high-quality SMPL pose and shape parameters for in-the-wild videos. It contains a total of 58 minutes of motion data, distributed over 81 indoor and outdoor sequences and 10 participants, which are collected with body-worn, wireless electromagnetic (EM) sensors and a hand-held iPhone. Samples in EMDB are significantly different from commonly used datasets and are suitable for verifying the generalization of TAR. Following the split protocol in \cite{kaufmann2023emdb}, we evaluate TAR on EMDB 1 in Table \ref{tab:emdb}. Our TAR surpasses previous image-based methods on all metrics by a large margin. The results verify the generalization ability of TAR.

\begin{table}[htb]
\caption{Evaluation of state-of-the-art image-based methods and TAR on EMDB 1.}
\label{tab:emdb}
\resizebox{\columnwidth}{!}{%
\renewcommand\arraystretch{1.5}
\begin{tabular}{@{}c|ccccccc@{}}
\toprule[1.5pt]
Method    & MPJPE{[}mm{]}      & PA-MPJPE{[}mm{]}   & MVE{[}mm{]}         & MVE-PA{[}mm{]}     & MPJAE{[}deg{]}     & MPJAE-PA{[}deg{]}  & Jitter            \\ \midrule
PyMAF\cite{zhang2021pymaf}     & 131.1$\pm$54.9         & 82.9$\pm$38.2          & 160.0$\pm$64.5          & 98.1$\pm$44.4          & 28.5$\pm$12.5          & 25.7$\pm$10.1          & 81.8$\pm$25.6         \\
PARE\cite{kocabas2021pare}      & 113.9$\pm$49.5         & 72.2$\pm$33.9          & 133.2$\pm$57.4          & 85.4$\pm$39.1          & 24.7$\pm$9.8           & 22.4$\pm$8.8           & 75.1$\pm$22.5         \\
CLIFF\cite{li2022cliff}     & 103.1$\pm$43.7         & 68.8$\pm$33.8          & 122.9$\pm$49.5          & 81.3$\pm$37.9          & 23.1$\pm$9.9           & 21.6$\pm$8.6           & 55.5$\pm$17.9         \\
HybrIK\cite{li2021hybrik}    & 103.0$\pm$44.3         & 65.6$\pm$33.3          & 122.2$\pm$50.5          & 80.4$\pm$39.1          & 24.5$\pm$11.3          & 23.1$\pm$11.1          & 49.2$\pm$18.5         \\
TAR(Ours) & \textbf{89.4}$\pm$48.0 & \textbf{58.9}$\pm$35.0 & \textbf{102.7}$\pm$53.2 & \textbf{69.3}$\pm$38.9 & \textbf{21.5}$\pm$15.2 & \textbf{21.4}$\pm$17.4 & \textbf{12.0}$\pm$6.6 \\ \bottomrule[1.5pt]
\end{tabular}
}
\end{table}

\begin{table}[htb]
\centering
\caption{Evaluation of ReFit\cite{wang2023refit}, SmoothNet\cite{zeng2022smoothnet} and TAR on 3DPW dataset. $\ddagger$ represents our implementation.}
\label{tab:smoothnet}
\renewcommand\arraystretch{1.3}
\begin{tabular}{@{}ccccc@{}}
\toprule[1.5pt]
Method                                     & MPJPE$\downarrow$         & PA-MPJPE$\downarrow$      & PVE$\downarrow$           & ACCEL$\downarrow$         \\ \midrule
\multicolumn{1}{c|}{ReFit\cite{wang2023refit}$\ddagger$}                 & 65.7          & 41.2          & 75.1          & 25.9          \\
\multicolumn{1}{c|}{ReFit\cite{wang2023refit}$\ddagger$+SmoothNet\cite{zeng2022smoothnet}(T=8)}  & \textbf{62.7} & 41.3 & \textbf{73.6} & 12.1          \\
\multicolumn{1}{c|}{ReFit\cite{wang2023refit}$\ddagger$+SmoothNet\cite{zeng2022smoothnet}(T=16)} & \textbf{62.7}          & 41.5          & \textbf{73.6}          & 10.3 \\
TAR(Ours)                                  & \textbf{62.7}        & \textbf{40.6}          & 74.4          & \textbf{7.7}           \\ \bottomrule[1.5pt]
\end{tabular}
\end{table}

\subsubsection{Comparison with SmoothNet}
SmoothNet\cite{zeng2022smoothnet} shows remarkable performance in improving accuracy and smoothness of 2D/3D keypoints and human shape recovery models. It is method-agnostic and and gains consistent improvements on public 3DPW dataset for image-based methods\cite{kolotouros2019learning}\cite{kocabas2021pare}. Since TAR is built on top of stronger base model than generally used SPIN\cite{kolotouros2019learning}, we compare our method with SmoothNet and a recent state-of-the-art ReFit\cite{wang2023refit} to verify the performance in terms of accuracy and temporal-consistency of TAR. As shown in Table, our TAR achieves comparable 3D accuracy and surpassing smoothness with SmoothNet with larger window size. The evaluation further proves the effectiveness of TAR. 

\begin{figure}[htb]
\centering
\includegraphics[width=\textwidth]{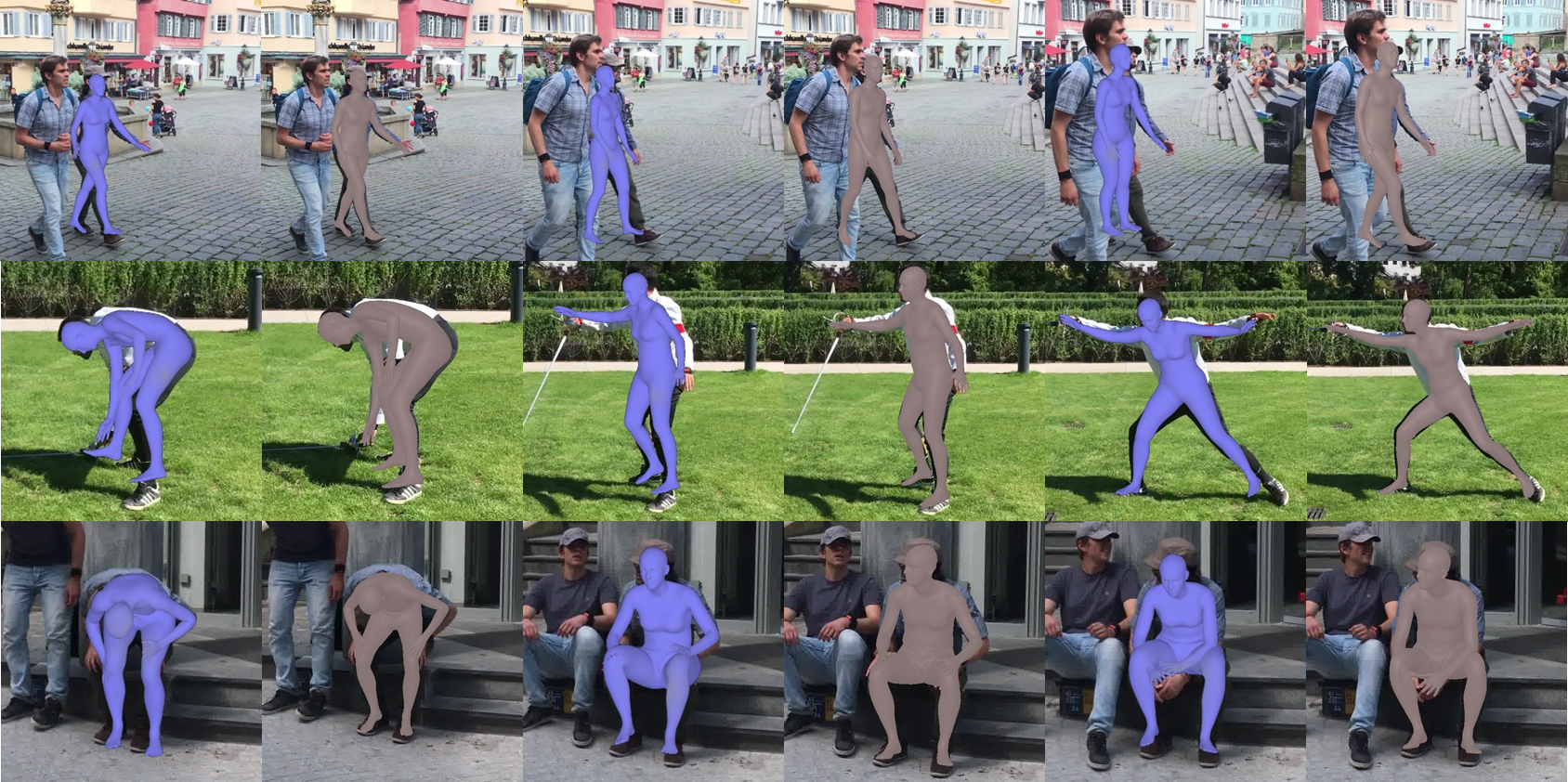}
\caption{Visual comparison between GLoT (purple meshes) and our TAR (brown meshes) on the challenging 3DPW dataset. Our method can generate more plausible and image-aligned mesh results than GLoT.}
\label{fig:qual-pw3d}
\end{figure}

\begin{figure}[htb]
\centering
\includegraphics[width=\textwidth]{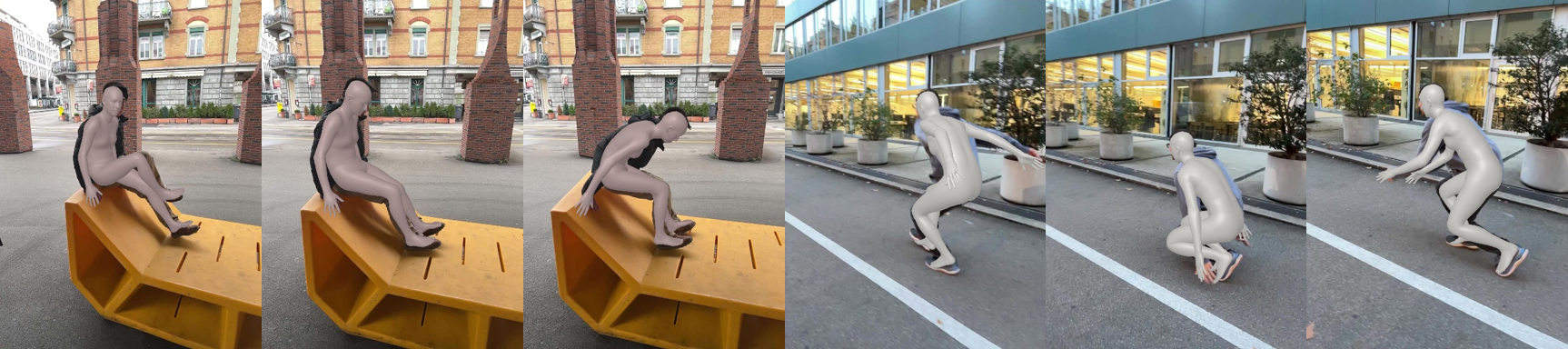}
\caption{Qualitative results of TAR on challenging EMDB dataset.}
\label{fig:qual-emdb}
\end{figure}

\begin{figure}[htb]
\centering
\includegraphics[width=\textwidth]{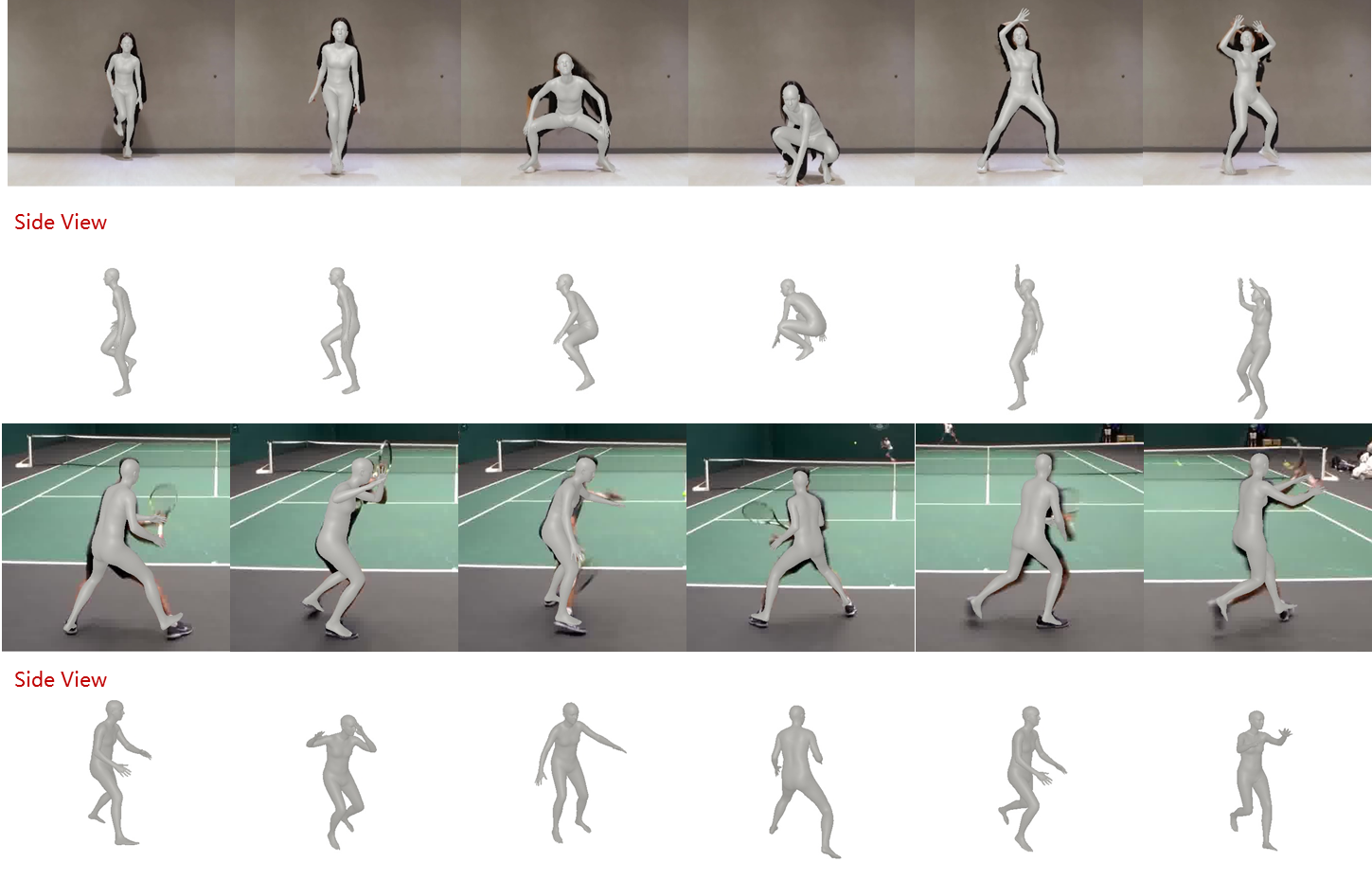}
\caption{Qualitative results of TAR on challenging internet videos.}
\label{fig:qual-wild}
\end{figure}

\subsection{Qualitative Evaluation}

\subsubsection{Visual Comparison with GLoT\cite{shen2023global}.}
Figure \ref{fig:qual-pw3d} shows the qualitative comparison between the previous state-of-the-art method GLoT and our TAR on the in-the-wild 3DPW dataset. Our method produces results with better 3D accuracy and better alignment with 2D images. Our method can achieve more reasonable results when an occlusion occurs (Row 1) by exploiting both temporal and spatial relations in a video sequence. 

\subsubsection{Qualitative results on EMDB and internet videos.}
Figure \ref{fig:qual-emdb} shows the qualitative results on challenging EMDB dataste. Videos in EMDB dataset are recorded with a hand-held Iphone, which frequently violates the weakly projection assumption. Our TAR can generate smooth, accurate and well-aligned human meshes on EMDB videos. In Figure \ref{fig:qual-wild}, we validate our method on challenging internet videos and provide rendered meshes of an side viewpoint. The results demonstrate that our TAR can handle challenging real scenarios.

\section{Conclusions}
This paper proposes a novel Temporal-Aware Refining Network (TAR)
for video-based 3D human pose and shape recovery, which collaboratively
exploits both temporal global and local features of videos with Global/Local Temporal Encoder and Recurrent Refinement Module. Extensive experiments demonstrate that TAR achieves state-of-the-art performance on three widely used datasets. Furthermore, TAR shows remarkable generalizing capability on challenging in-the-wild EMDB dataset and internet videos. We hope that our approach will spark further research in video-based 3D human pose and shape recovery.
\subsection{Limitations.}
Our method produces degraded result when the human body is highly occluded. The future work will explore the physical constraints and motion prior to further improve the performance. TAR is slightly inferior to SOTA video-based methods on inter-frame metric, which is our future optimization direction.

\clearpage

\bibliographystyle{unsrtnat}
\bibliography{references}

\clearpage

\appendix
\title{Supplementary Material}
\beginsupplement

\section{Additional Quantitative Results}
\subsection{Comparison with more image-based methods.}
Table \ref{tab:supp-3dpw} compares our TAR with image-based and video-based methods on the 3DPW dataset. All methods are trained on the training set of 3DPW and report the best result using either HRNet\cite{wang2020deep} or ResNet50\cite{he2016deep} backbones. Image-based methods focus on per-frame accuracy with advanced network design to extract image features, while video-based methods focus on temporal-consistency and use a pretrained backbone provided by \cite{kolotouros2019learning}. Therefore, image-based methods generally achieve better 3D accuracy than their video-based counterparts. In contrast, our TAR designs a new temporal architecture to exploit both global and local image features to achieve both accuracy and temporal-consistency. Our method outperforms image-based methods in metrics of MPJPE, PA-MPJPE and PVE, and achieve comparable ACCEL metric with state-of-the-art video-based methods. The results demonstrate the superiority and effectiveness of our designs for 3D human motion estimation.

\subsection{Trained wtih extra data.}
We trained TAR with BEDLAM\cite{black2023bedlam}, a recently launched synthetic dataset of 3D human motions, to further verify the performance. As shown in Table \ref{tab:supp-bedlam}, TAR achieves better performance on 3DPW dataset with BEDLAM, demonstrating the generalization of our method.

\begin{table}[htb]
\centering
\caption{Comparison to the state-of-the-arts image-based and video-based methods on 3DPW.  All methods are trained on the 3DPW training set and report the best results.}
\label{tab:supp-3dpw}
\begin{tabular}{@{}lcccc@{}}
\toprule
\multicolumn{2}{c|}{\multirow{2}{*}{Method}} & \multicolumn{3}{c}{3DPW} \\
\multicolumn{2}{c|}{} & MPJPE & PA-MPJPE & PVE \\ \midrule
\multirow{9}{*}{image-based} & \multicolumn{1}{c|}{PyMAF\cite{zhang2023pymaf}} & 74.2 & 45.3 & 87.0 \\
 & \multicolumn{1}{c|}{PARE\cite{kocabas2021pare}} & 74.5 & 46.5 & 88.6 \\
 & \multicolumn{1}{c|}{HybrIK\cite{li2023hybrik}} & 72.9 & 41.8 & 88.6 \\
 & \multicolumn{1}{c|}{NIKI\cite{li2023niki}} & 71.3 & 40.6 & 86.6 \\
 & \multicolumn{1}{c|}{CLIFF\cite{li2022cliff}} & 73.5 & 44.5 &  \\
 & \multicolumn{1}{c|}{ReFit\cite{wang2023refit}} & 65.8 & 41.0 & - \\
 & \multicolumn{1}{c|}{VisDB\cite{yao2022learning}} & 73.5 & 44.9 & 85.5 \\
 & \multicolumn{1}{c|}{Virtual Marker\cite{ma20233d}} & 67.5 & 41.3 & 77.9 \\
 & \multicolumn{1}{c|}{FastMETRO\cite{cho2022cross}} & 73.5 & 44.6 & 84.1 \\ \midrule
\multirow{8}{*}{video-based} & \multicolumn{1}{c|}{VIBE\cite{kocabas2020vibe}} & 91.9 & 57.6 & 99.1 \\
 & \multicolumn{1}{c|}{MEVA\cite{luo20203d}} & 86.9 & 54.7 & - \\
 & \multicolumn{1}{c|}{TCMR\cite{choi2021beyond}} & 86.5 & 52.7 & 102.9 \\
 & \multicolumn{1}{c|}{MAED\cite{wan2021encoder}} & 79.1 & 45.7 & 92.6 \\
 & \multicolumn{1}{c|}{MPS-Net\cite{wei2022capturing}} & 84.3 & 52.1 & 99.7 \\
 & \multicolumn{1}{c|}{INT\cite{yang2023capturing}} & 75.6 & 42.0 & 87.9 \\
 & \multicolumn{1}{c|}{GLoT\cite{shen2023global}} & 80.7 & 50.6 & 96.3 \\
 & TAR(Ours) & \textbf{62.7} & \textbf{40.6} & \textbf{74.4} \\ \bottomrule
\end{tabular}%
\end{table}

\begin{table}[htb]
\centering
\caption{Evaluation: with additional synthetic data from BEDLAM.}
\label{tab:supp-bedlam}
\renewcommand\arraystretch{1.3}
\begin{tabular}{c|cccc}
\toprule[1.5pt]
Method       & MPJPE$\downarrow$         & PA-MPJPE$\downarrow$      & PVE$\downarrow$           & ACCEL$\downarrow$        \\ \midrule
TAR          & 62.7          & 40.6          & 74.4          & \textbf{7.7 }         \\
TAR+BEDLAM\cite{black2023bedlam} & \textbf{60.4} & \textbf{38.8} & \textbf{70.7} & 7.8          \\ \bottomrule[1.5pt]
\end{tabular}
\end{table}

\begin{figure}[htb]
\centering
\includegraphics[width=\textwidth]{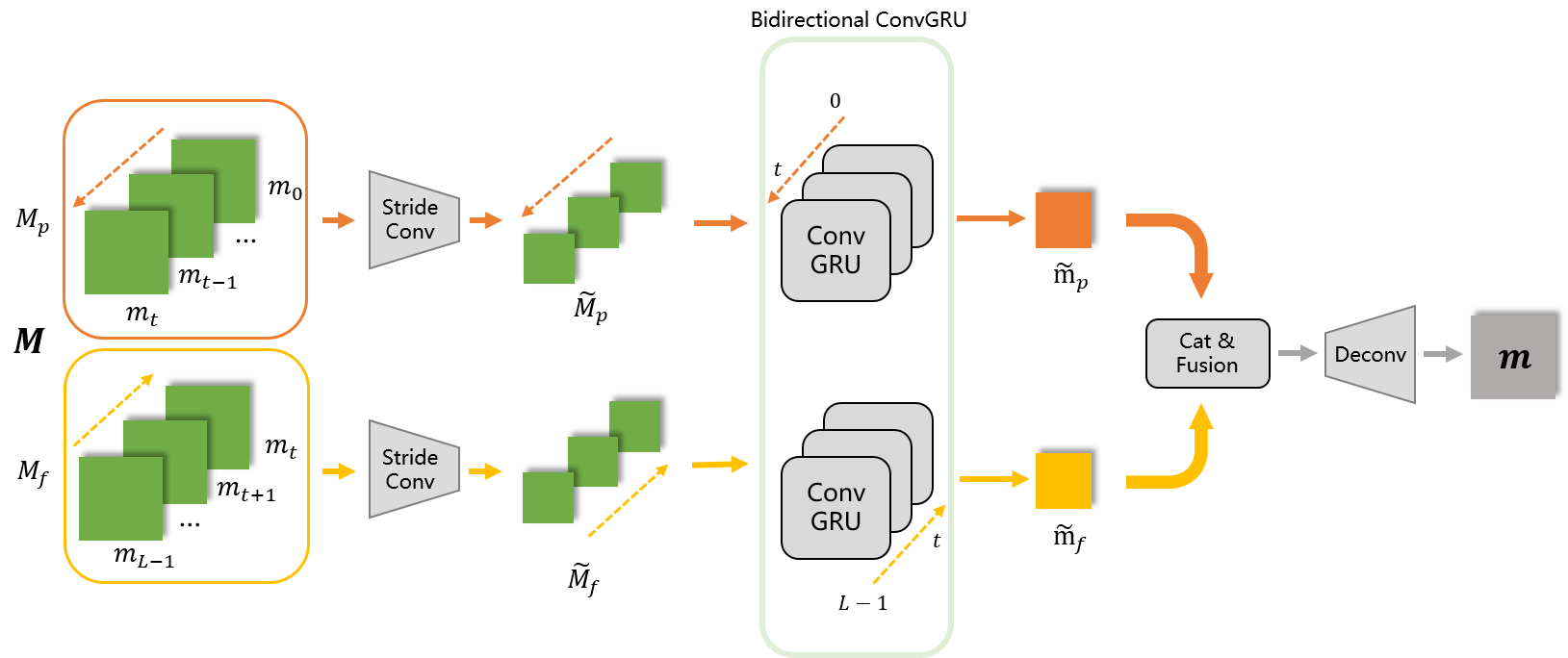}
\caption{Local Temporal Encoder with downsampled hidden maps. }
\label{fig:ds_lte}
\end{figure}

\section{Additional Ablation Study}
\subsection{Size of hidden maps in ConvGRU}
TAR use ConvGRU to model the temporal information of high-resolution local feature maps extracted by CNNs\cite{he2016deep}\cite{wang2020deep}. Specifically, when the input image crops have shape of $256\times 256$, the local feature maps will be $64\times 64$. Although this design is effective for modeling local feature maps, it also brings high computation burdens. In order to reduce the amount of computation and improve the real-time practicality of the model, we provide a variant structure for Local Temporal Encoder as shown in Figure \ref{fig:ds_lte}. Specifically, we first downsample the local feature maps via a group of strided convolutional layers to reduce the size of hidden maps of ConvGRU. Then the encoded hidden maps $\mathbf{\tilde{m}_f}$ and $\mathbf{\tilde{m}_p}$ are concatenated and fed to deconvolutional layers to restore the feature size. We conduct different downsampling rates to evaluate the impact of hidden feature size of ConvGRU. As shown in Table \ref{tab:supp-size}, when the size of hidden maps decreases, the accuracy and smoothness will slightly decrease. In return, the computational complexity is significant reduced. Even with small size of hidden maps (e.g.,$S$=8), TAR still outperforms most state-of-the-art image-based and video-based methods in accuracy and achieves satisfying temporal-consistency.

\begin{table}[htb]
\centering
\caption{Evaluation of different sizes of ConvGRU hidden maps. $S$ is the size of hidden maps in ConvGRU.}
\label{tab:supp-size}
\renewcommand\arraystretch{1.3}
\begin{tabular}{@{}c|cccc|cc@{}}
\toprule[1.5pt]
Method & MPJPE$\downarrow$ & PA-MPJPE$\downarrow$ & PVE$\downarrow$ & ACCE$\downarrow$L & GFLOPs$\downarrow$ & Params(M)$\downarrow$ \\ \midrule
S=64 & 62.7 & 40.6 & 74.4 & 7.7 & 184.6 & 25.0 \\
S=32 & 65.1 & 43.4 & 77.8 & 8.1 & 72.4 & 29.9 \\
S=16 & 67.3 & 44.3 & 81.0 & 8.5 & 43.5 & 34.3 \\
S=8 & 66.9 & 43.1 & 80.1 & 8.4 & 36.3 & 38.5 \\ \bottomrule[1.5pt]
\end{tabular}%
\end{table}

\section{Additional Qualitative Results}

\begin{figure}[htb]
\centering
\includegraphics[width=\textwidth]{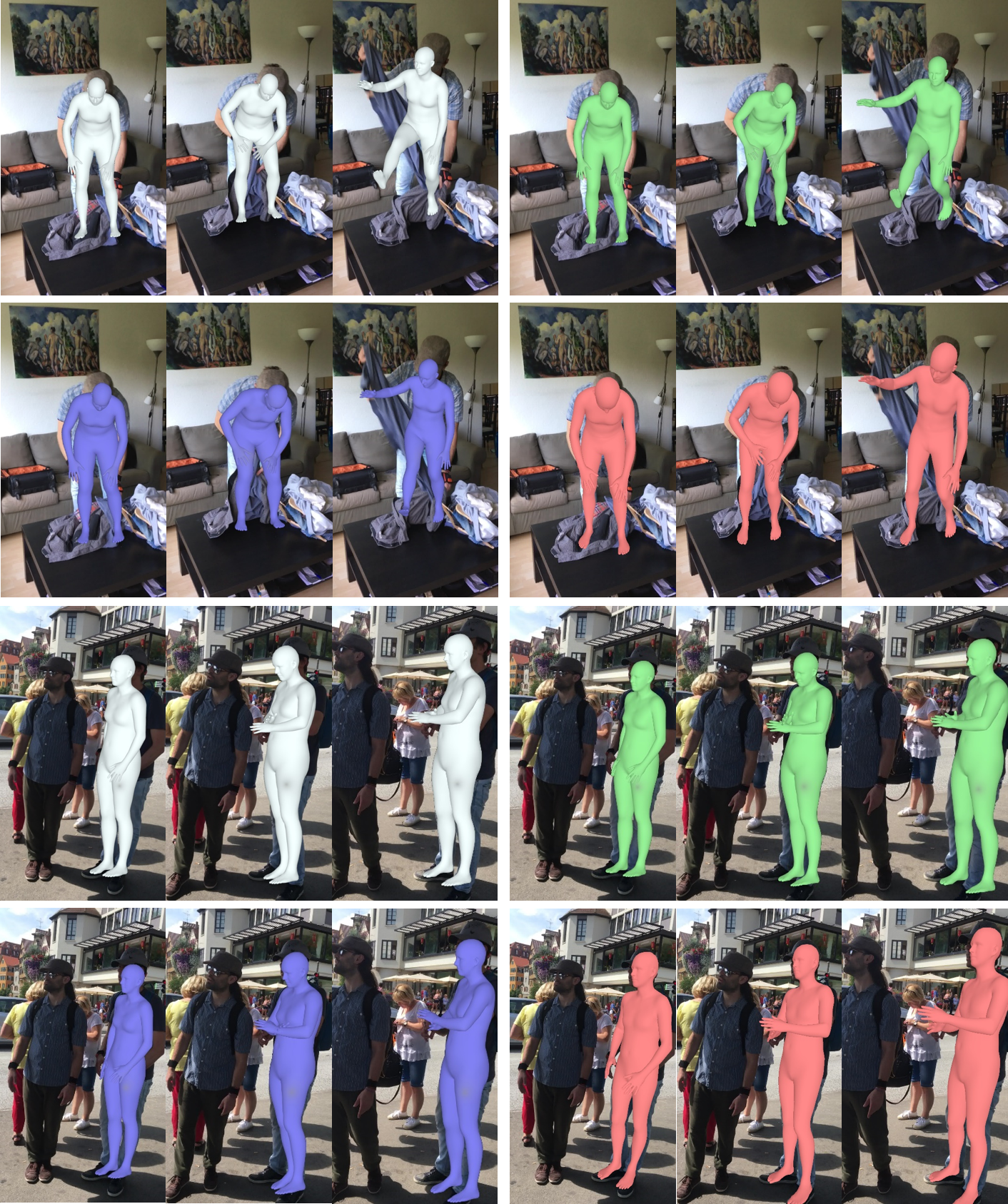}
\caption{Visual comparison between \textcolor[rgb]{0.9,0.9,0.9}{TCMR}, \textcolor[rgb]{0.3,0.9,0.3}{MPS-Net}, \textcolor[rgb]{0.3,0.3,0.9}{GLoT} and \textcolor[rgb]{0.9,0.3,0.3}{TAR}. }
\label{fig:supp-3dpw}
\end{figure}

\begin{figure}[htb]
\centering
\includegraphics[width=\textwidth]{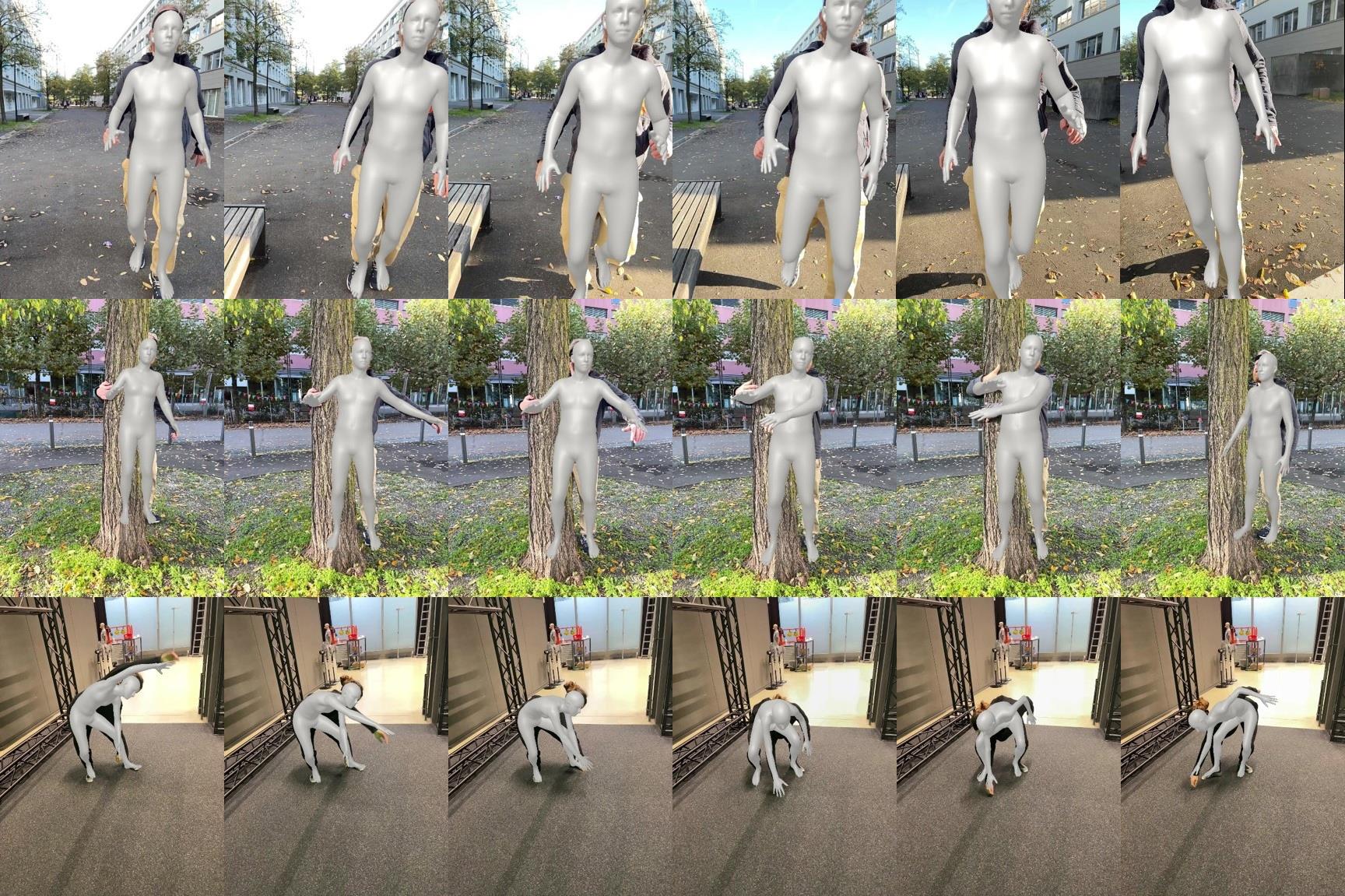}
\caption{Visual evaluation on EMDB\cite{kaufmann2023emdb} samples.}
\label{fig:supp-emdb}
\end{figure}

\begin{figure}[htb]
\centering
\includegraphics[width=\textwidth]{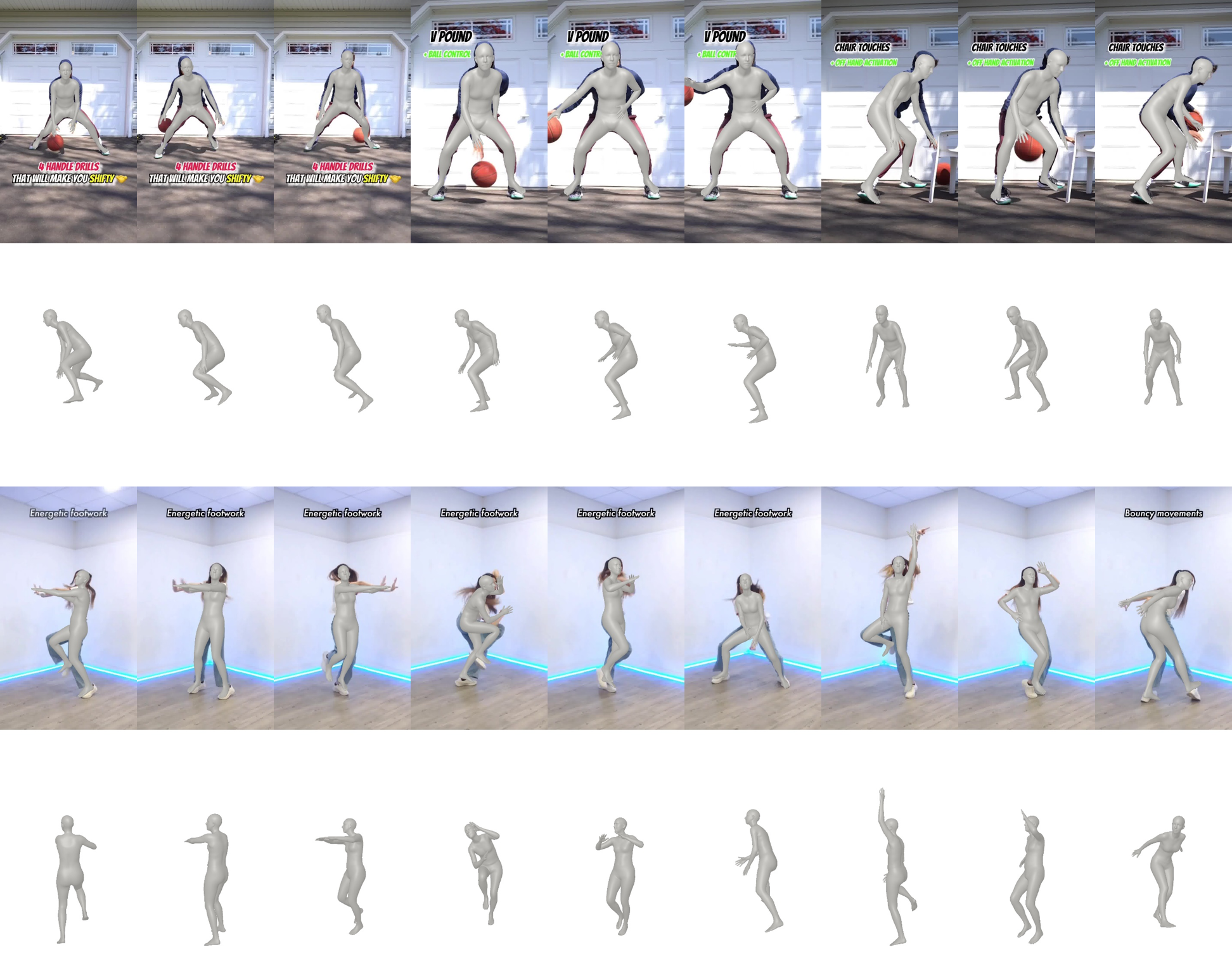}
\caption{Visual evaluation on internet videos.}
\label{fig:supp-wild}
\end{figure}

We show more comparison results with other methods (TCMR\cite{choi2021beyond}, MPS-Net\cite{wei2022capturing} and GLoT\cite{shen2023global}) in Figure \ref{fig:supp-3dpw}. Figure \ref{{fig:supp-emdb}} gives more test results of EMDB, which contain truncation, occlusion, and challenging poses. Figure\ref{fig:supp-wild} shows more challenging internet videos, including dancing and sports scenarios.

\end{document}